\documentclass{article}

\usepackage{PRIMEarxiv}

\usepackage[utf8]{inputenc} 
\usepackage[T1]{fontenc}    
\usepackage{hyperref}       
\usepackage{url}            
\usepackage{booktabs}       
\usepackage{amsfonts}       
\usepackage{nicefrac}       
\usepackage{microtype}      
\usepackage{lipsum}
\usepackage{fancyhdr}       
\usepackage{graphicx}       
\graphicspath{{media/}}     
\usepackage{amsthm}

\newtheorem{definition}{Definition}

\usepackage{subfigure}
\usepackage{multirow}
\usepackage[linesnumbered,ruled,vlined]{algorithm2e}
\usepackage{balance}
\usepackage{amsmath}

\title{Towards Fast and Stable Federated Learning: Confronting Heterogeneity via Knowledge Anchor
} 

\author{
  Jinqian Chen, Jihua Zhu\thanks{Corresponding Author} \\
  Xi'an Jiaotong University \\
  Xi'an, China\\
  \texttt{chenjinqian@stu.xjtu.edu.cn, zhujh@xjtu.edu.cn} \\
   \And
  Qinghai Zheng \\
  Fuzhou University \\
  Fuzhou, China\\
  \texttt{zhengqinghai@fzu.edu.cn}}

\begin{document}
\maketitle
\begin{abstract}
Federated learning encounters a critical challenge of data heterogeneity, adversely affecting the performance and convergence of the federated model. Various approaches have been proposed to address this issue, yet their effectiveness is still limited. Recent studies have revealed that the federated model suffers severe forgetting in local training, leading to global forgetting and performance degradation. Although the analysis provides valuable insights, a comprehensive understanding of the vulnerable classes and their impact factors is yet to be established. In this paper, we aim to bridge this gap by systematically analyzing the forgetting degree of each class during local training across different communication rounds. Our observations are: (1) Both missing and non-dominant classes suffer similar severe forgetting during local training, while dominant classes show improvement in performance. (2) When dynamically reducing the sample size of a dominant class, catastrophic forgetting occurs abruptly when the proportion of its samples is below a certain threshold, indicating that the local model struggles to leverage a few samples of a specific class effectively to prevent forgetting. Motivated by these findings, we propose a novel and straightforward algorithm called Federated Knowledge Anchor (FedKA). Assuming that all clients have a single shared sample for each class, the knowledge anchor is constructed before each local training stage by extracting shared samples for missing classes and randomly selecting one sample per class for non-dominant classes. The knowledge anchor is then utilized to correct the gradient of each mini-batch towards the direction of preserving the knowledge of the missing and non-dominant classes. Extensive experimental results demonstrate that our proposed FedKA achieves fast and stable convergence, significantly improving accuracy on popular benchmarks.
\end{abstract}


\section{Introduction}
Data serves as the foundation of artificial intelligence. Despite the pervasive influence of informatization across various domains \cite{levinson2011towards, miotto2016deep}, the availability of high-quality data remains limited. Such inadequacy may not necessarily stem from a natural scarcity, but rather from restricted access or collection due to security and privacy concerns. The isolation of data significantly impedes the real-world application of artificial intelligence \cite{yang2019federated}.

\begin{figure}[t!]
    \centering
    \subfigure[Class-wise test accuracy]{\includegraphics[width=0.30\textwidth ]{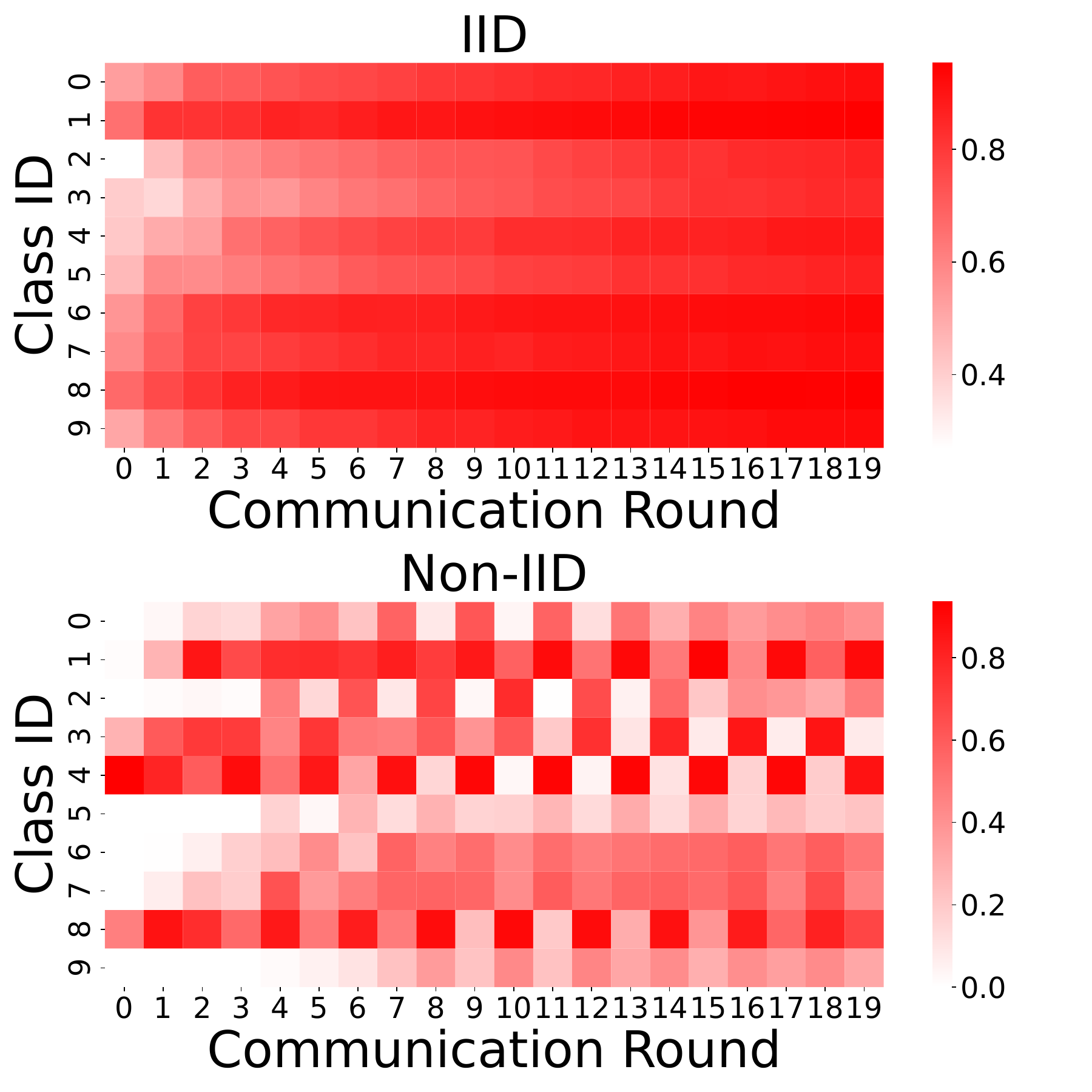}}
    \subfigure[Global test accuracy]{\includegraphics[width=0.30\textwidth]{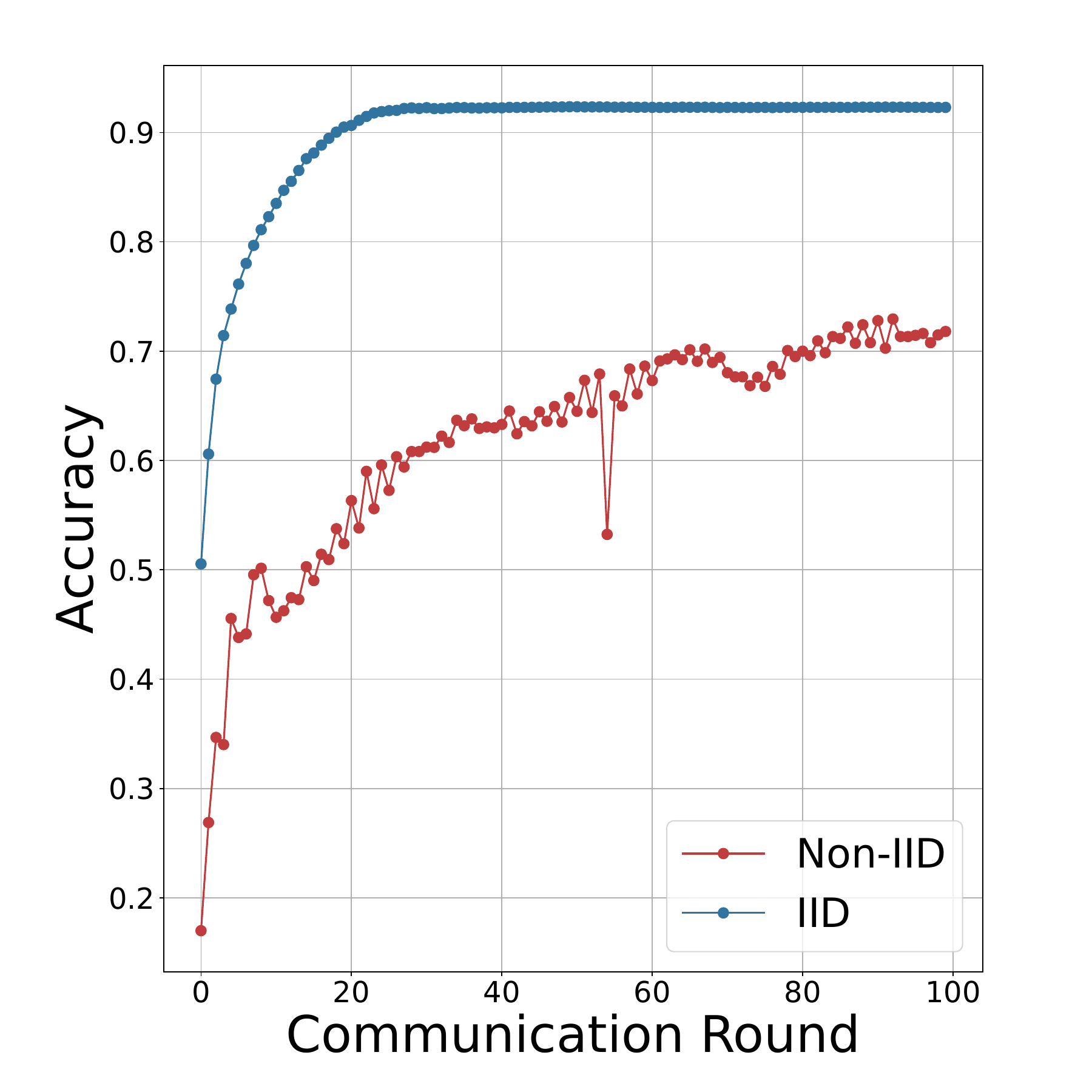}}
    \caption{Forgetting problem in federated learning on Non-IID data. Although class-wise accuracy remains consistent in IID settings, it can dramatically fluctuate when heterogeneous data is involved. This indicates that federated models suffer from forgetting on heterogeneous data, which hinders convergence and constrains aggregated model performance.}
    \label{fig1_1}
\end{figure}

To address these issues, McMahan et al. \cite{mcmahan2017communication} introduced the concept of federated learning and developed the classic algorithm, FedAvg. Federated learning enables clients to train a model in a distributed manner without sharing their private data. In each communication round of FedAvg, the central server initially distributes the current global model to each client. Subsequently, each client trains the model using their local data for several epochs and uploads the updated parameter to the central server. The uploaded parameters are then aggregated through averaging to generate the global model for the next round. Federated learning is an efficient and privacy-preserving approach that has been successfully applied to various practical scenarios, such as smart city \cite{smartcity, li2020review}, smart health care \cite{smart-health-care, dayan2021federated}, etc.

However, as federated learning is increasingly adopted, several critical issues \cite{li2020federated, bagdasaryan2020backdoor} have emerged, including the challenge of heterogeneity. Traditional deep learning methods assume that data is independent and identically distributed (IID), which ensures an upper bound on generalization error and provides a strong theoretical basis for machine learning. However, this assumption does not always hold in practical federated learning scenarios, as clients operate in diverse environments and possess heterogeneous data (also referred to as Non-IID data below) \cite{li2020federated}.

Heterogeneous data in federated learning can be divided into several types: attribute skew, label skew, quantity skew, and their combinations \cite{zhu2021federatednonIID, FLExperimentalStudy}. Attribute skew refers to scenarios in which the data attributes differ among clients (e.g. clients possess text and clients possess image), which is always the research topic of vertical federated learning.
However, label skew focuses on differences in label distributions between clients (e.g. clients possess more cat samples and clients possess more tiger samples). Quantity skew refers to the imbalance of data sample numbers between clients, which always co-occurs with label skew. In this paper, we focus on the most common label skew in horizontal federated learning. As noted in \cite{li2020federated, li2020federatedprox}, heterogeneity significantly hinders model performance and leads to severe convergence problems.

Numerous methods \cite{li2020federatedprox, li2021modelmoon} have been proposed to alleviate the impact of heterogeneous data in federated learning. However, these existing methods have not adequately addressed the issue. Studies have revealed that some methods that claim to be better than the baseline FedAvg actually perform worse in severe heterogeneous scenarios. Recently, Lee et al. \cite{lee2022preservationntd} provided a systematic analysis of federated learning on heterogeneous data from the perspective of forgetting. They found that clients are prone to forgetting the knowledge of out-local data, which results in forgetting in the global model. However, their study formulates forgetting from the perspective of in-local and out-local data, while the situation of class-wise forgetting remains unclear.

In this paper, we present extensive experimental results demonstrating that catastrophic forgetting only affects missing classes and classes with a small number of samples, referred to as non-dominant classes. In contrast, the performance of dominant classes does not degrade and even improves during local training. By dynamically adjusting the number of samples of a specific class, we further illustrate that catastrophic forgetting of a particular class suddenly appears when the number of samples falls below a certain threshold. Our findings indicate that the local model cannot effectively utilize the few samples possessed by non-dominant classes to combat forgetting during training, leading to severe performance degradation similar to that observed in missing classes. Such catastrophic forgetting in missing and dominant classes impedes convergence and hinders the improvement of the federated model's performance.

\begin{figure}[t!]
    \centering
    \subfigure[Data distribution visualization]{\includegraphics[width=0.30\textwidth ]{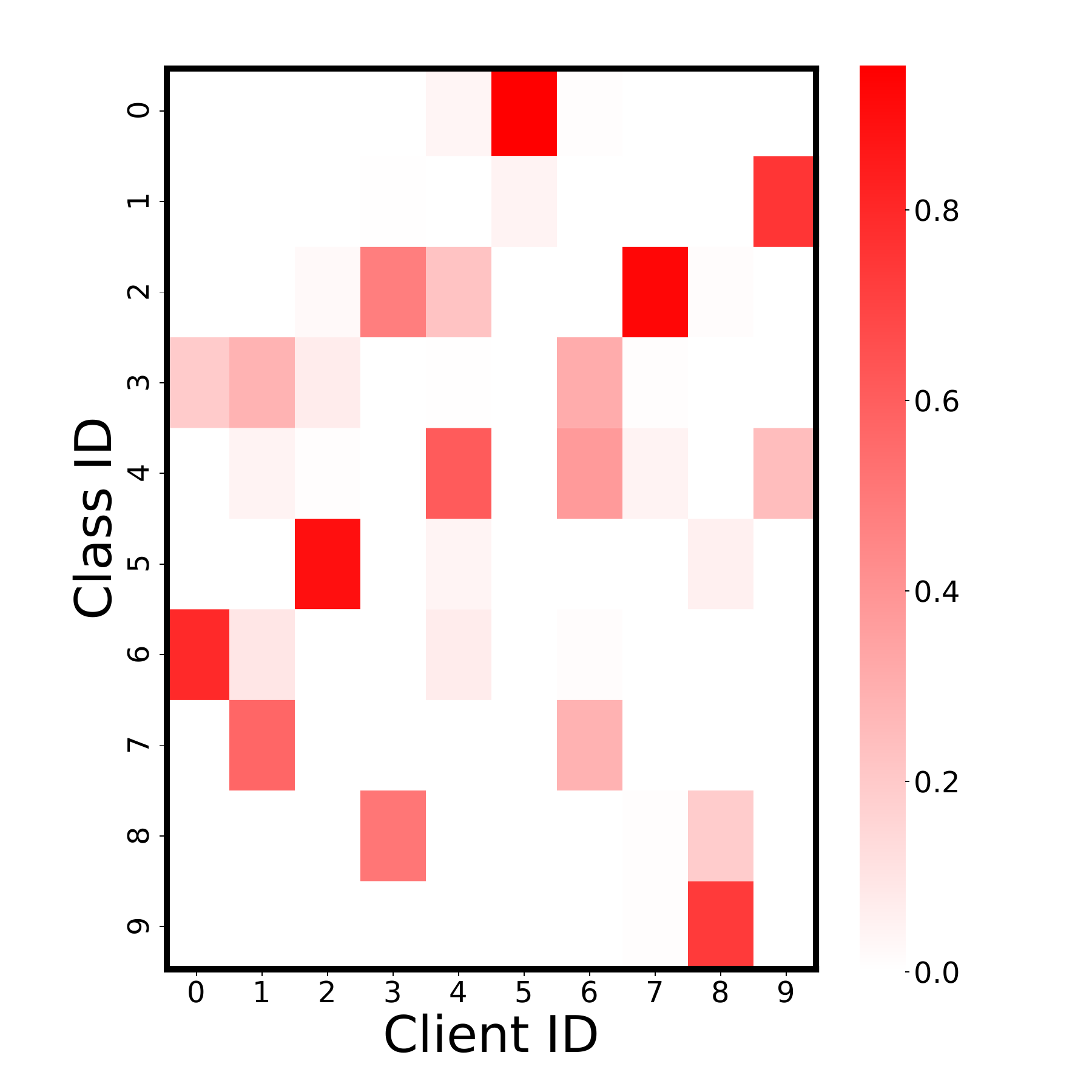}} 
    \subfigure[Forgetting degree of client 3]{\includegraphics[width=0.30\textwidth]{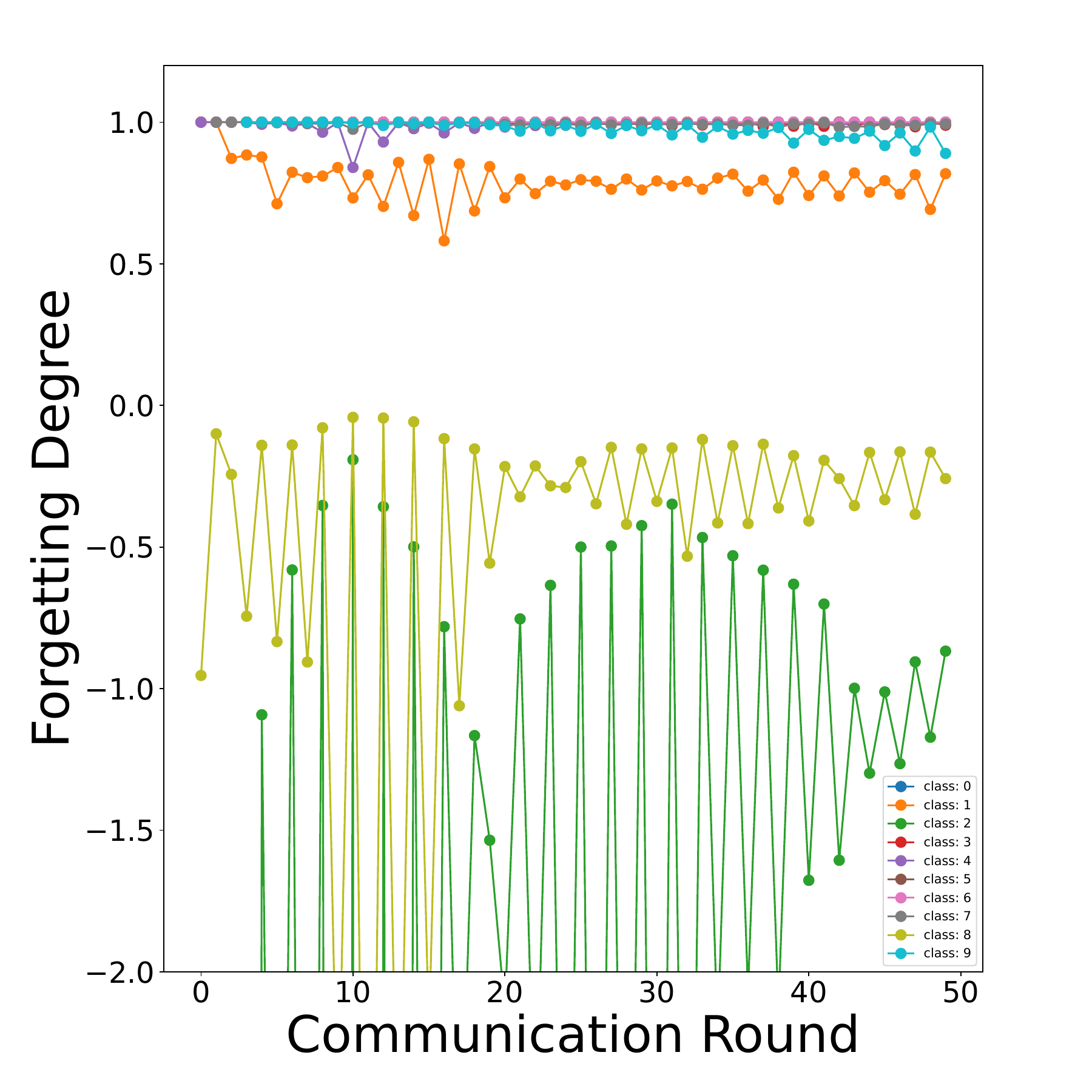}}
    \caption{Class-wise forgetting in local models: (a) Data distribution visualization of the heterogeneous Cifar10. (b) Forgetting degree of different classes in the local model of client 3. Dominant classes in client 3, namely class 2 and 8, exhibit a negative forgetting degree indicating performance improvement in local training. However, the other non-dominant and missing classes suffer severe forgetting in local training. }
    \label{fig3_1}
\end{figure}

Motivated by the findings of catastrophic forgetting in missing and non-dominant classes, we propose a novel algorithm, Federated Knowledge Anchor (FedKA), to mitigate this issue in heterogeneous data. Specifically, we first construct a shared dataset containing only one shared sample per class. During each local training stage of a specific client, we use the shared sample to compensate for the missing classes. For each non-dominant class, we randomly select one sample from the possessed data. The two sample sets representing the missing classes and the non-dominant classes together form what we call "knowledge anchor". We further utilize the knowledge anchor in each mini-batch training to minimize the L2 distance between the discarded logits of the knowledge anchor predicted by the current global model and the updated local model. This encourages the model to preserve the knowledge of the missing and non-dominant classes. We conduct extensive experiments to demonstrate the effectiveness of our proposed FedKA. The experimental results demonstrate that FedKA achieves state-of-the-art performance with high accuracy, fast convergence, and stability.

In summary, our contributions are as follows:

1. We provide a further analysis of catastrophic forgetting in federated learning, demonstrating that catastrophic forgetting only affects missing and non-dominant classes during local training, while the performance of dominant classes is improved. Furthermore, we show that the few samples of non-dominant classes possessed by the clients are not fully utilized to combat forgetting.

2. We propose an efficient algorithm, called Federated Knowledge Anchor (FedKA), to address the heterogeneous issues regarding catastrophic forgetting. FedKA utilizes our proposed knowledge anchor to correct the direction of gradients in the mini-batch and mitigate catastrophic forgetting in non-dominant and missing classes.

3. We conduct extensive experiments to evaluate our proposed algorithm. The results demonstrate that our FedKA achieves state-of-the-art accuracy, convergence, and robustness.

\section{Background and Related Work}

\subsection{Problem Setup}
Consider a horizontal federated learning scenario with $N$ clients in a $K$-class image classification problem. Each client $c_i$ possesses its private dataset $\mathcal{D}_i = \{(\boldsymbol{x}_j, y_j)\}_{j=1}^{n_i} = \{\mathcal{D}_i^1, ... , \mathcal{D}_i^K\}$, where $\mathcal{D}_i^k$ is the client $i$'s sample set of class $k$ and could be empty. The goal of the federated learning is to collaboratively train a model $f_{\boldsymbol{\theta}}$ on the global dataset $\mathcal{D} = \mathcal{D}_1 \cup ... \cup \mathcal{D}_N $ without the leakage of private data in $\mathcal{R}$ communication rounds. At each communication round $r$, the server sends the current global model parameter ${\boldsymbol{\theta}^r_g}$ to each participating client. After receiving the global parameters, the client $c_i$ trains this model $f_{\boldsymbol{\theta^r_i}}$ using its private dataset $\mathcal{D}_i$ for $E$ epochs to get the updated local model parameter $\boldsymbol{\theta}_i^r$. The updated parameters are subsequently sent to the central server and aggregated to obtain the updated global model parameter $\boldsymbol{\theta}^{r+1}_g$ for the next round. Note that the classification model $f_{\boldsymbol{\theta}}$ mentioned in this paper does not contain the final Softmax operation.

\subsection{Related Work}
\textbf{Federated Learning (FL)} is a decentralized training paradigm for deep learning, allowing clients to train a model collaboratively without uploading their private data. The most commonly used algorithm for federated learning is FedAvg \cite{mcmahan2017communication}. However, FedAvg suffers from severe performance degradation when dealing with heterogeneous data. To address this issue, numerous methods have been proposed from different perspectives \cite{zhu2021federatednonIID, FLExperimentalStudy}. One of the most common strategies is client drift mitigation. The idea behind client drift is that the optimal solution for the objective function under local data may not be the optimal solution for global data. Some examples of these methods include FedProx \cite{li2020federatedprox}, SCAFFOLD \cite{karimireddy2020scaffold}, MOON \cite{li2021modelmoon}, and FedDyn \cite{FedDyn}, etc. Another set of methods focuses on the aggregation schema, including FedAvgM \cite{hsu2019measuringFedAvgM}, FedNova \cite{wang2020tacklingfednova}, etc. FedAvgM uses momentum updates for gradient aggregation, while FedNova normalizes local updates before aggregation. Besides the aforementioned generic federated learning methods which aim to train a single model to generalize on all clients' data, personalized federated learning is also a feasible solution to tackle the heterogeneity. The personalized federated learning allows each client to possess its own model and utilize the global aggregated information to assist its training \cite{deng2020adaptive, tan2022towards, pFedMe}. SOTA personalized federated learning methods adopt various strategies to personalize their local model, including fine-tuning global model \cite{Per-FedAvg}, splitting and personalizing the projection heads  \cite{FedRep}, personalized aggregation \cite{FedAMP, FedFOMO, FedALA}, additional personalized models \cite{pFedMe, ditto}, etc. Though personalized federated models demonstrate great performance improvement in heterogeneous data, the generic performance and the generalization ability of these models are limited. Recent studies also reveal that several generic federated methods perform even worse than FedAvg on severe heterogeneous data \cite{luo2021no}. All of these call for a better generic federated framework to tackle the heterogeneity problem.

\textbf{Forgetting in FL.} Catastrophic forgetting (CF) is a long-study question in continual learning but a novel perspective to address issues caused by the data heterogeneity in federated learning. CF in continual learning and CF in federated learning both refer to significant performance degradation in neural networks over time. However, they differ in several aspects. In continual learning, CF occurs at the task level due to the addition of new domains or tasks \cite{de2021continual}. In federated learning, CF happens at the class level because of the label distribution shift when the global model is trained on clients' data \cite{lee2022preservationntd}. Furthermore, CF in continual learning is monolithic, affecting the entire model, whereas, in federated learning, CF exists independently in each client, jointly influencing the global model during aggregation.
While the concept was first introduced in FedCurv \cite{shoham2019overcoming}, a comprehensive analysis of the relationship between forgetting and the impact of heterogeneous data was lacking until the introduction of FedNTD \cite{lee2022preservationntd}. FedNTD highlights that clients tend to forget the knowledge of out-of-distribution data, leading to model forgetting and hindering performance on heterogeneous data. However, the analysis overlooks the influence of label distribution between classes on forgetting, particularly with regard to non-dominant and missing classes.

\section{Forgetting on Non-dominant and Missing Classes.}
To further investigate the differences in catastrophic forgetting among different classes and its influencing factors, we conduct extensive experiments on the Cifar10 dataset \cite{cifar10}. We employ a four-layer convolutional neural network, referred to as t-CNN, as described in the Appendix. Similar to \cite{li2021modelmoon}, we split the Cifar10 dataset into 10 heterogeneous clients using the Dirichlet distribution $\textbf{Dir}(\alpha)$, where we set $\alpha = 0.1$ to induce significant data heterogeneity. Each local training stage consisted of 10 epochs.

\subsection{Class-wise Forgetting in Local Models}

To investigate the classes most susceptible to catastrophic forgetting in heterogeneous data, we measure the class-wise forgetting degree for each local training stage of clients. The forgetting degree $\tau_{i, r}^{k}$ of class $k$ for client $i$ in communication round $r$ is defined as:
\begin{equation}
   \tau_{i, r}^{k} = \frac{\textbf{acc}(\boldsymbol{\theta}^r_g;k) - \textbf{acc}(\boldsymbol{\theta}_i^r; k)}{\textbf{acc}(\boldsymbol{\theta}^r_g;k) + \xi} 
\end{equation}, where $\textbf{acc}(\boldsymbol{\theta}; k)$ is the test accuracy of the model with parameter $\boldsymbol{\theta}$ on class $k$, and $\xi$ is a small positive constant to maintain numerical stability. The forgetting degree $\tau$ ranges from $- \infty$ to $1$. A positive value of $\tau_{i,r}^{k}$ indicates that client $i$ has forgotten knowledge of class $k$ after local training in communication round $r$, while a negative value indicates that the knowledge of class $k$ has improved for client $i$. The closer the value of $\tau$ is to 1, the more severe the forgetting.

\begin{definition}
Consider a local dataset $\{(\boldsymbol{x}_j, y_j)\}_{j=1}^{n_i}$ of client $c_i$ with $n_i$ data samples in a $K$-class classification problem. Given the certain threshold $\gamma$, the missing classes $C_m^i$, non-dominant classes $C_n^i$, dominant classes $C_d^i$ of client $c_i$ are:

$$C_m^i = \{k|\sum_{j=1}^{n_i}{\mathbb{I}(y_j = k)} = 0, k<K, k \in \mathbb{Z}\}$$
$$C_n^i = \{k|0 < \frac{1}{n_i}\sum_{j=1}^{n_i}{\mathbb{I}(y_j = k)} < \gamma , k<K, k \in \mathbb{Z}\}$$
$$C_d^i = \{k|\frac{1}{n_i}\sum_{j=1}^{n_i}{\mathbb{I}(y_j = k)} \geq \gamma , k<K, k \in \mathbb{Z}\}$$

\end{definition}

We train the t-CNN using FedAvg for 50 communication rounds and calculated the class-wise forgetting degree for each client across the communication rounds. To provide a more intuitive conclusion, we also visualize the data distribution. The results are presented in Fig. \ref{fig3_1}.

Our findings suggest that the forgetting of each class is significantly influenced by the number of samples in that class. The forgetting degree $\tau$ of the non-dominant classes $k \in C_n$ or the missing classes $k \in C_m$ is always greater than 0 and close to 1, whereas the forgetting degree $\tau$ of the dominant classes $k \in C_d$ is consistently negative. These observations suggest that the non-dominant or missing classes experience significant forgetting during local training, while dominant classes exhibit improvement in performance.

\begin{figure}[t!]
    \centering

    \subfigure[Reduction on the non-dominant class]{\includegraphics[width=0.35\textwidth ]{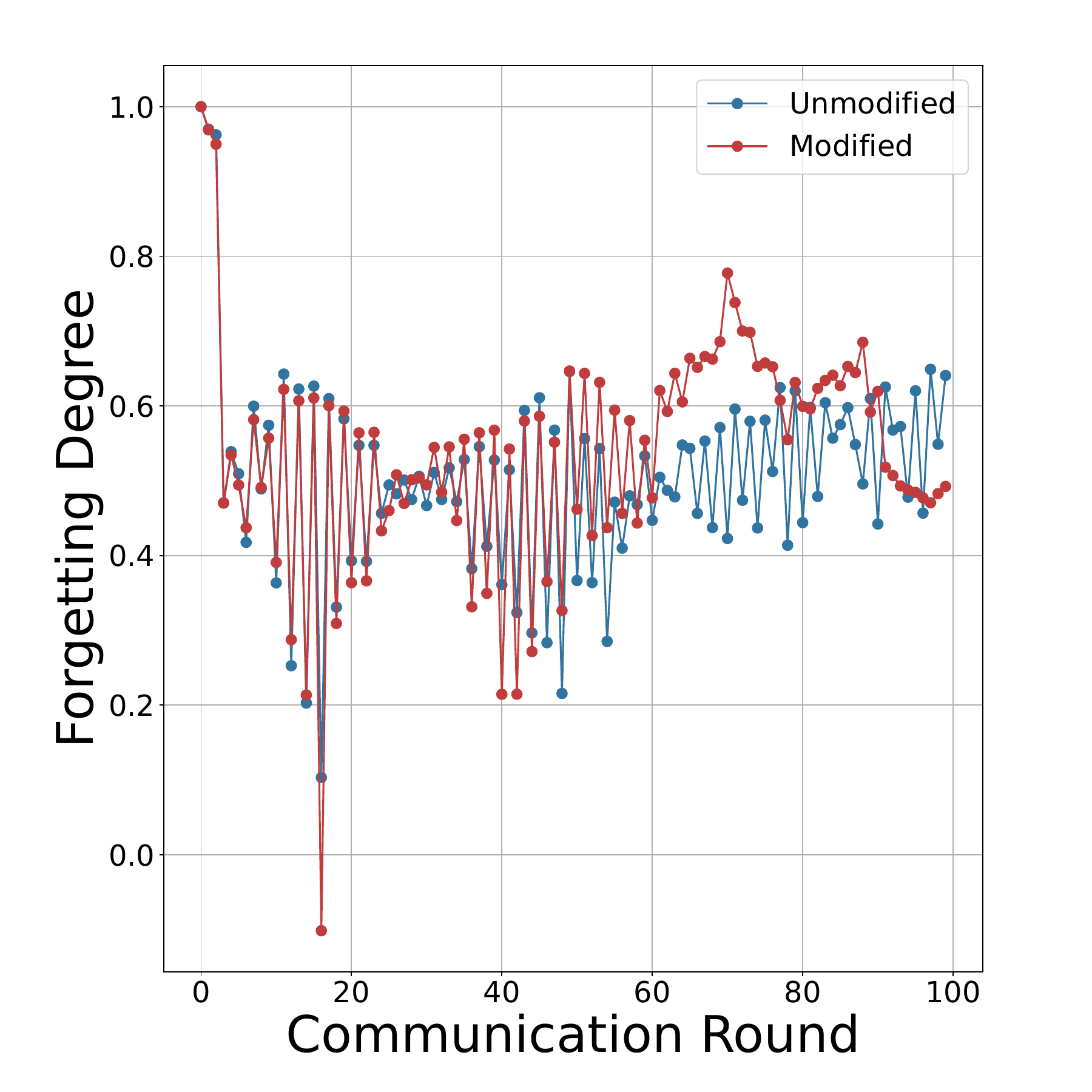}}
    \subfigure[Reduction on the dominant class]{\includegraphics[width=0.35\textwidth]{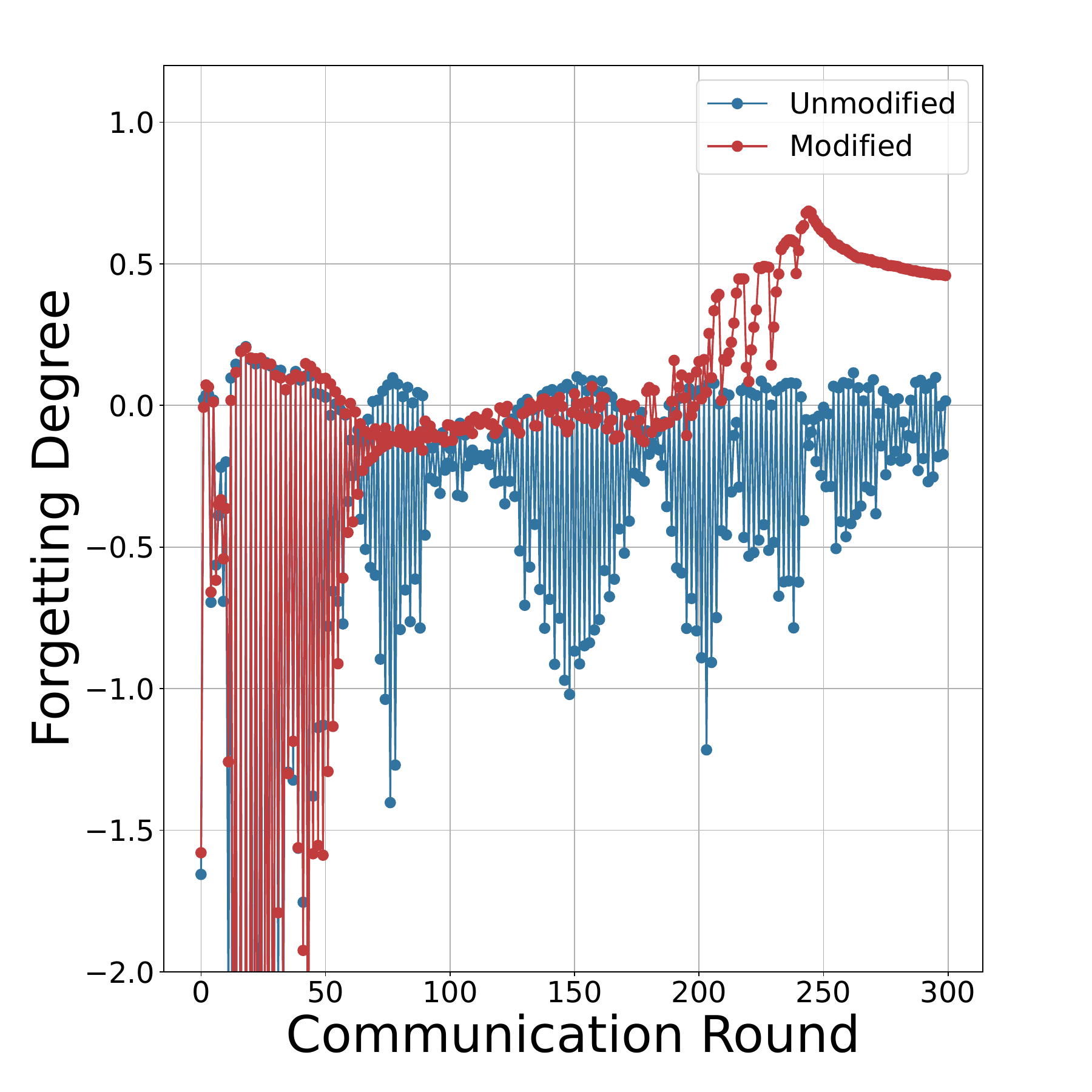}}
    \caption{Forgetting degrees of dominant and non-dominant classes under various label distributions. We dynamically discard several samples of the specified class to change its label distribution. In (a), we discard samples of class $1$, representing non-dominant classes, at communication rounds 50, 60, and 70, reducing the number of samples to 10, 5, and 0, respectively. In (b), we discard samples of class $3$, representing dominant classes, after 50 communication rounds. The results demonstrate that reducing the number of samples has a trivial impact on the forgetting degree of non-dominant classes, indicating that clients cannot make full use of samples from these classes. However, reducing the number of samples from dominant classes significantly reduces the performance benefits obtained in local training, and catastrophic forgetting occurs when the number of samples falls below a certain threshold.}
    \label{fig3_2}
\end{figure}
\subsection{Forgetting under Various Label Distribution}

After analyzing the above observation, we investigate how forgetting varies with different label distributions. One naive assumption is that the more samples a specific class has, the less severe the forgetting will be. However, the truth remains unclear. To validate this assumption, we adopt a strategy that dynamically reduces the sample number of a specific class $k$ and observes the changes in its forgetting degree $\tau^k$.

We apply this strategy to class $1$ and class $3$ respectively on the client $0$, representing the sample number reduction on non-dominant and dominant classes. In our simulations, client $0$ has nearly 5000 data samples, with the number of samples of class $3$ close to $1000$, while the number of samples of class $1$ is only about $10$. For class $3$ representing dominant classes, we first train the federated model without any modification for 50 communication rounds to get a relatively stable state of the federated framework. After 50 communication rounds, we discard the data samples of class $3$, which account for about $1\%$ of the total number of samples possessed by client $0$, every ten rounds, so that the proportion of samples of class 3 in the total samples is gradually reduced from $20\%$ to $1\%$. For the non-dominant class $1$, we discard the sample number to $10, 5, 0$ respectively at the communication round $50, 60, 70$. We display the results in Fig. \ref{fig3_2}.

Strikingly, the results indicate a sharp contrast in the effect of sample reduction on dominant and non-dominant classes. In the case of non-dominant classes, the forgetting rate slightly increases following the reduction in the 50th communication round but subsequently decreases to an even lower rate that is similar to the unmodified level. In contrast, for the dominant class, sample reduction leads to a significant decrease in knowledge acquisition during local training, resulting in a sharp increase in the forgetting rate. Furthermore, when the proportion of the dominant class is reduced to 5\%, catastrophic forgetting suddenly occurs.

Based on our findings, we conclude that local models in federated learning cannot adequately utilize limited samples to prevent forgetting, resulting in similar forgetting rates in non-dominant classes and catastrophic forgetting in dominant classes.

\begin{figure*}[tbp]
    \centering
    \includegraphics[width=0.95\textwidth]{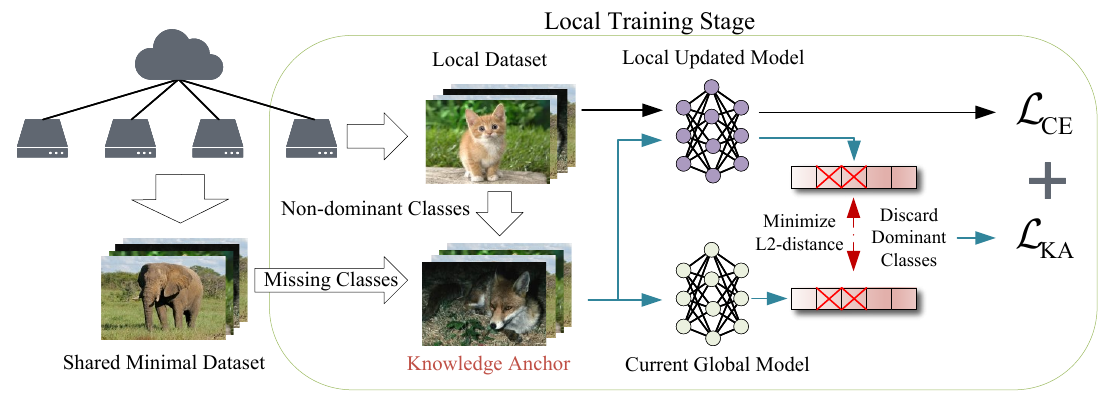}
    \caption{The overall framework of Federated Knowledge Anchor (FedKA). FedKA mandates all clients to build a minimal shared dataset comprising one sample for each class before training. Then, in each local training stage of different communication rounds, clients construct their own knowledge anchors based on their data distribution. The knowledge anchor comprises one sample for each dominant and missing class and is utilized to adjust the mini-batch gradients in the direction of preserving knowledge on non-dominant and missing classes.}
    \label{overall}
\end{figure*}

\section{FedKA: Federated Knowledge Anchor}
In this section, we propose Federated Knowledge Anchor (FedKA). The core idea of FedKA is to utilize a minimal subset of the global data distribution as anchor points to force the local model to preserve the knowledge of non-dominant and missing classes. Through the construction and the utilization of the knowledge anchor, the local model can make full use of the few samples possessed by the non-dominant classes and mitigate the catastrophic forgetting caused by missing samples of specific classes.

\begin{definition}
Consider a client $c_i$ with its dominant, non-dominant, missing classes $C_d^i$, $C_n^i$ and $C_m^i$ in federated learning. Assume all clients share a minimal dataset $S$ which possesses one single sample per class for the $K$-class classification problem and $|\mathcal{S}| = K$. The knowledge anchor of client $c_i$ is:

\begin{equation}
    \mathcal{T}_i = \{\psi(k, \mathcal{D}_i, \mathcal{S})| k \in C_m^i \cup C_n^i\}
\end{equation}, where $$
\psi(k, \mathcal{D}_i, \mathcal{S}) =\left \{ \begin{array}{lcl}
(\boldsymbol{x}, y) \sim D_i^k & & {k \in C_n^i}\\
(\boldsymbol{x}, y) \sim \{(\boldsymbol{x}, y) | (\boldsymbol{x}, y) \in \mathcal{S}, y = k \} & & {k \in C_m^i}\\
\end{array} \right.$$
\end{definition}

Knowledge anchor $\mathcal{T}_i$ is the minimal sample subset that compensates for the missing classes and non-dominant classes in the client $c_i$. Note that $|\mathcal{T}_i| = |C_m^i|+|C_n^i|$.

We now describe our method Federated Knowledge Anchor (FedKA). Before training, FedKA requires all clients to create a shared dataset $S$ which contains only one sample for each class. Such an assumption does not violate the principles of privacy and is quite easy to satisfy as client number $N$ is always greater than $K$ in federated learning. For each communication round, every participating client $c_i$ will first construct its knowledge anchor $\mathcal{T}_i^r$. Then the knowledge anchor $\mathcal{T}_i^r$ is utilized to correct the local gradient of each mini-batch towards the direction of preserving the knowledge of the missing and non-dominant classes. Specifically, we borrow the Not-True-Distillation (NTD) idea from \cite{lee2022preservationntd} and extend it to all non-dominant and missing classes. In each mini-batch update, we add an additional regularization term to minimize the L2 distance between the knowledge anchor's logits (the last-layer outputs before the Softmax operation) of the current global model and the updated local model. Note that the logits represented for dominant classes are discarded. The loss function of client $c_i$ in local training is:
\begin{equation}
\mathcal{L} = \mathcal{L}_{CE} + \beta \cdot \mathcal{L}_{KA}\left(\mathcal{T}_i^r, f_{\boldsymbol{\theta}^r_g}, f_{\boldsymbol{\theta}^r_i}\right)
\end{equation}, where $$
\mathcal{L}_{KA}\left(\mathcal{T}_i^r, f_{\boldsymbol{\theta}^r_g}, f_{\boldsymbol{\theta}^r_i}\right) = \frac{1}{\left|\mathcal{T}_i^r\right|}\left|\left|\Phi\left(f_{\boldsymbol{\theta}^r_g}\left(\mathcal{T}_i^r\right), C_d^i\right) - \Phi\left(f_{\boldsymbol{\theta}^r_i}\left(\mathcal{T}_i^r\right), C_d^i\right)\right|\right|^2
$$ and $\Phi$ is the discarding function that discards all logits of dominant classes in $C_d^i$. In classification tasks with a large number of classes, the proportion between local samples in a mini-batch and the sample number in the knowledge anchor $|\mathcal{T}_i|$ may become imbalanced. To address this issue, we adopt a sampling strategy to obtain a down-sampled knowledge anchor with a limited number of samples for mini-batch local training. We provide the full FedKA algorithm in Algorithm 1.

The underlying concept behind Federated Knowledge Anchor is that local training benefits dominant classes, but leads to catastrophic forgetting in non-dominant and missing classes. We aim for the knowledge anchor to act as anchor points in local knowledge preservation, which can alleviate the convergence problem and improve the model performance of federated learning on heterogeneous data. Although this approach is intuitive, theoretical analysis of the knowledge anchor is still challenging, and we leave it as a topic for future research.

\section{Experiments}
\subsection{The Setup}
In this section, we briefly introduce the experimental setup. Without the explicit declaration, we all conduct the experiments under the default setting mentioned below.

\textbf{ Datasets.} We conduct experiments on Cifar10 \cite{cifar10}, Cifar100 \cite{cifar10}, and Tiny-ImageNet \cite{le2015tiny}. Following \cite{li2021modelmoon, luo2021no}, we utilize Dirichlet distribution $\text{Dir} (\alpha)$ to generate Non-IID data among clients. The default $\alpha$ is set to 0.1 to simulate a strict Non-IID scenario. Such low values tend to result in the SOTA methods performing even worse than baseline FedAvg \cite{mcmahan2017communication} as indicated in \cite{luo2021no, FLExperimentalStudy}. And the default client number is set to 10.

\textbf{Comparing Methods.} We compare our FedKA with SOTA methods to show its superior performance, including FedAvg \cite{mcmahan2017communication}, FedProx \cite{li2020federatedprox}, MOON \cite{li2021modelmoon}, FedDyn \cite{FedDyn}, and FedNTD \cite{lee2022preservationntd}.

\textbf{Hyperparameters.} For fair comparison, all common hyperparameters are set to the same default value except for additional clarification. We set the communication rounds to 100 for Cifar10/Cifar100 and 20 for Tiny-ImageNet. The default local epoch number is set to 10. The batch size is set to 128, and the down-sampled number of the knowledge anchor is set to 10. For all methods involved in our experiments, we use the classic SGD optimizer with a learning rate of 0.01. The momentum of the SGD optimizer is set to 0.9, and the weight decay is set to 0.00001. All of these settings are common practice in \cite{li2021modelmoon, mcmahan2017communication}. To get the best performance of the comparing methods, we carefully tune the involved hyperparameters in each method. We choose the best $\mu$ in $\{0.1, 1, 5, 10\}$ and set $\tau$ to 0.5 for MOON as suggested in \cite{li2021modelmoon}. Following \cite{li2020federatedprox, li2021modelmoon}, we choose best $\mu$ from $\{0.001, 0.01, 0.1, 1\}$ for FedProx. For FedNTD, we also set $\beta = 1$ and $\tau = 1$ as instructed in its paper \cite{lee2022preservationntd}. The $\alpha$ used in FedDyn is chosen from $\{1, 0.1, 0.01, 0.001\}$ as instructed in its paper \cite{FedDyn}. The $\beta$ used in proposed FedKA is chosen from $\{0.5, 0.1, 0.01, 0.001\}$.

\textbf{Models.} We conduct experiments on two networks of different scales. For Cifar10, we train the t-CNN. For Cifar100 and Tiny-ImageNet, we train ResNet-50 \cite{he2016deep} for comparison. We modify the output projector of ResNet-50 to get the expected dimension.

\begin{algorithm}[t]
\caption{Federated Knowledge Anchor}
\begin{flushleft}
\item{
\KwIn{client number $N$, client set $\{c_1, ..., c_N\}$, dataset $\mathcal{D} = \mathcal{D}_1 \cup ... \cup \mathcal{D}_N$, local epochs $E$, communication round $\mathcal{R}$, limited knowledge anchor number $\mu$, model $f_{\boldsymbol{\theta}}$, loss weight $\beta$.}
}
\item{
    \For{each client $c_i$}{
  calculate its dominant, non-dominant, missing classes $C_d^i, C_n^i, C_m^i$.
}

  \textbf{construct} minimal shared dataset $\mathcal{S}$.

\textbf{Initialize} global model parameter. $\boldsymbol{\theta}^0_g$

    \For{each communication round $r = 1, ..., \mathcal{R}$}{
    sample the client set to get the participated client set $\mathcal{P}^r$.

    distribute the global model to each participated client:
    
    $\boldsymbol{\theta}^{r-1}_i \leftarrow \boldsymbol{\theta}^r_g$
    
    \For{each client $c_i$ in $\mathcal{P}^r$}{
        construct its knowledge anchor $\mathcal{T}_i^r$ for this round using: $\mathcal{T}_i^r = \{\psi(k, \mathcal{D}_i, \mathcal{S})| k \in C_m^i \cup C_n^i\}$.

        \If{$|\mathcal{T}_i^r| > \mu$}{
        Random discard $|\mathcal{T}_i^r| - \mu$ samples in knowledge anchor.
        
        }
        
        \For{each local step $e=1, ..., E$}{
        \For{each mini-batch B}{
            $\mathcal{L} = \mathcal{L}_{CE} + \beta \cdot \mathcal{L}_{KA}(\mathcal{T}_i^r, f_{\boldsymbol{\theta}^r_g}, f_{\boldsymbol{\theta}^r_i})$
        
            $\boldsymbol{\theta}_i^r \leftarrow \boldsymbol{\theta}_i^r - \nabla \mathcal{L}$
            
        }
        }
        upload $\boldsymbol{\theta}_i^r$ to server.
    }
    aggregate model parameters: $\boldsymbol{\theta}^{r+1}_{g} \leftarrow \sum_{c_i \in \mathcal{P}^r}\frac{|\mathcal{D}_i|}{|\mathcal{D}|} \boldsymbol{\theta}_i^r$
    }}
\item{
\KwOut{$\boldsymbol{\theta}_{g}^{\mathcal{R}}$}}

\end{flushleft}

\end{algorithm}

\begin{figure*}
    \centering
    \subfigure[Cifar10]{
    \includegraphics[width=0.29\textwidth ]{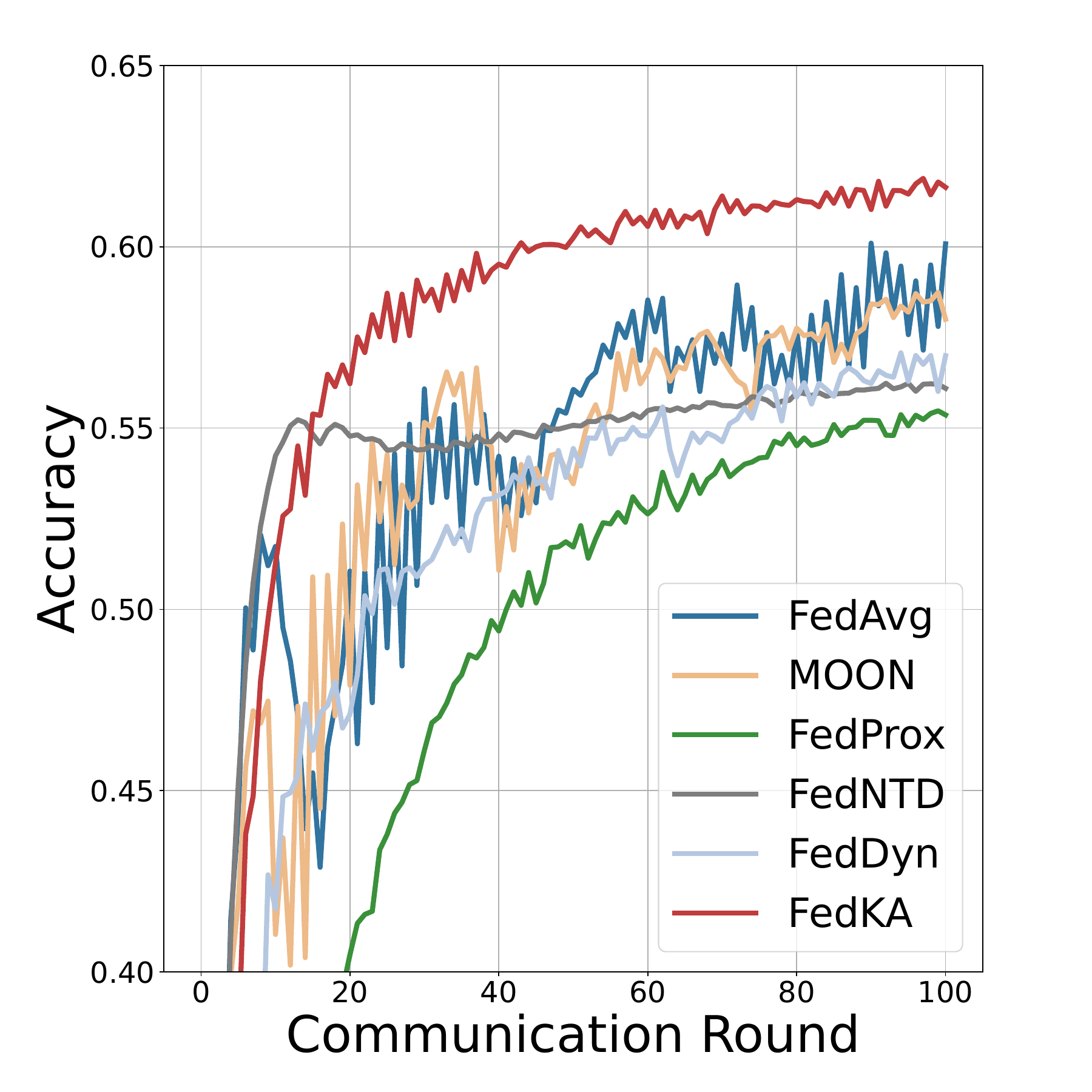}}
    \subfigure[Cifar100]{
    \includegraphics[width=0.29\textwidth ]{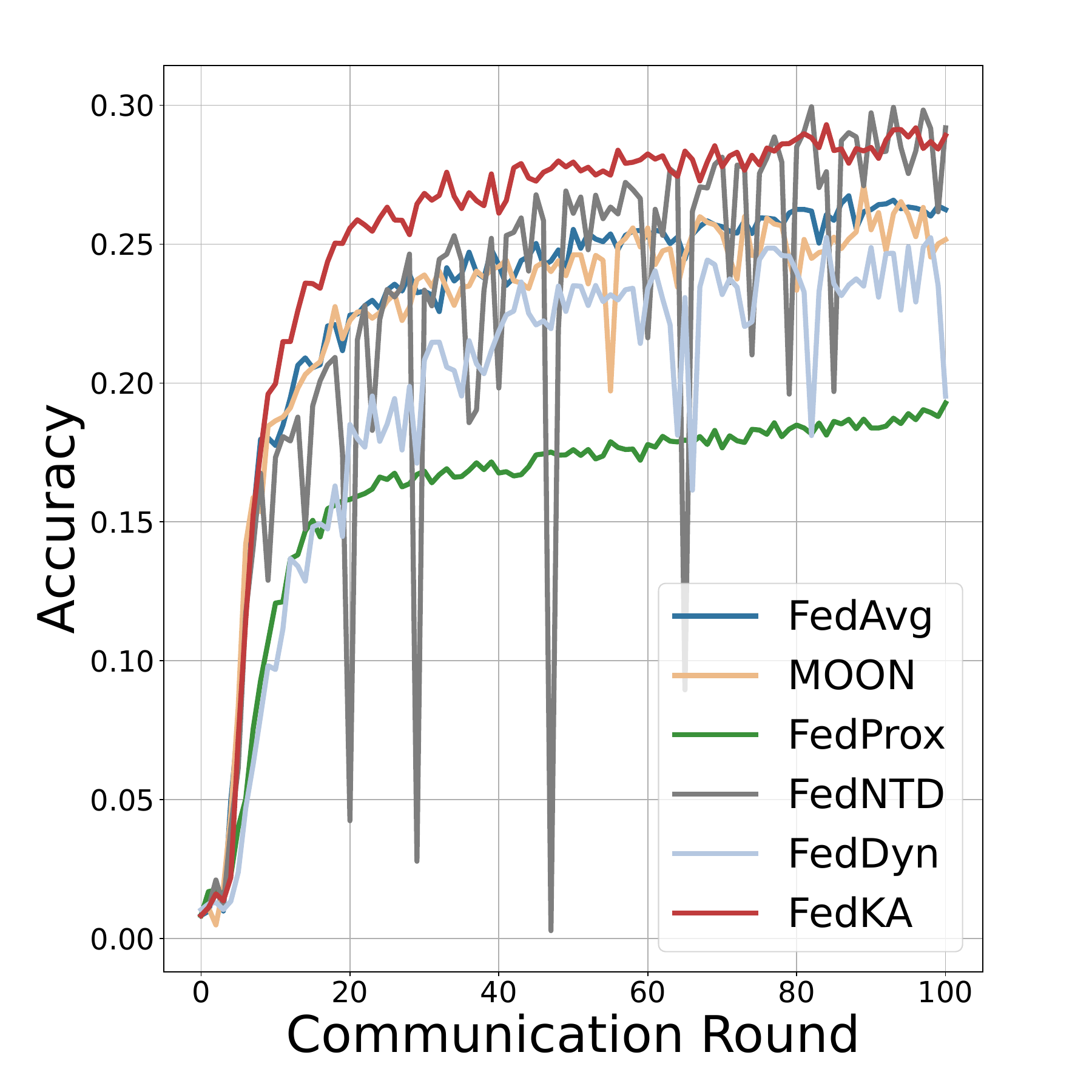}}
    \subfigure[Tiny-ImageNet]{
    \includegraphics[width=0.29\textwidth ]{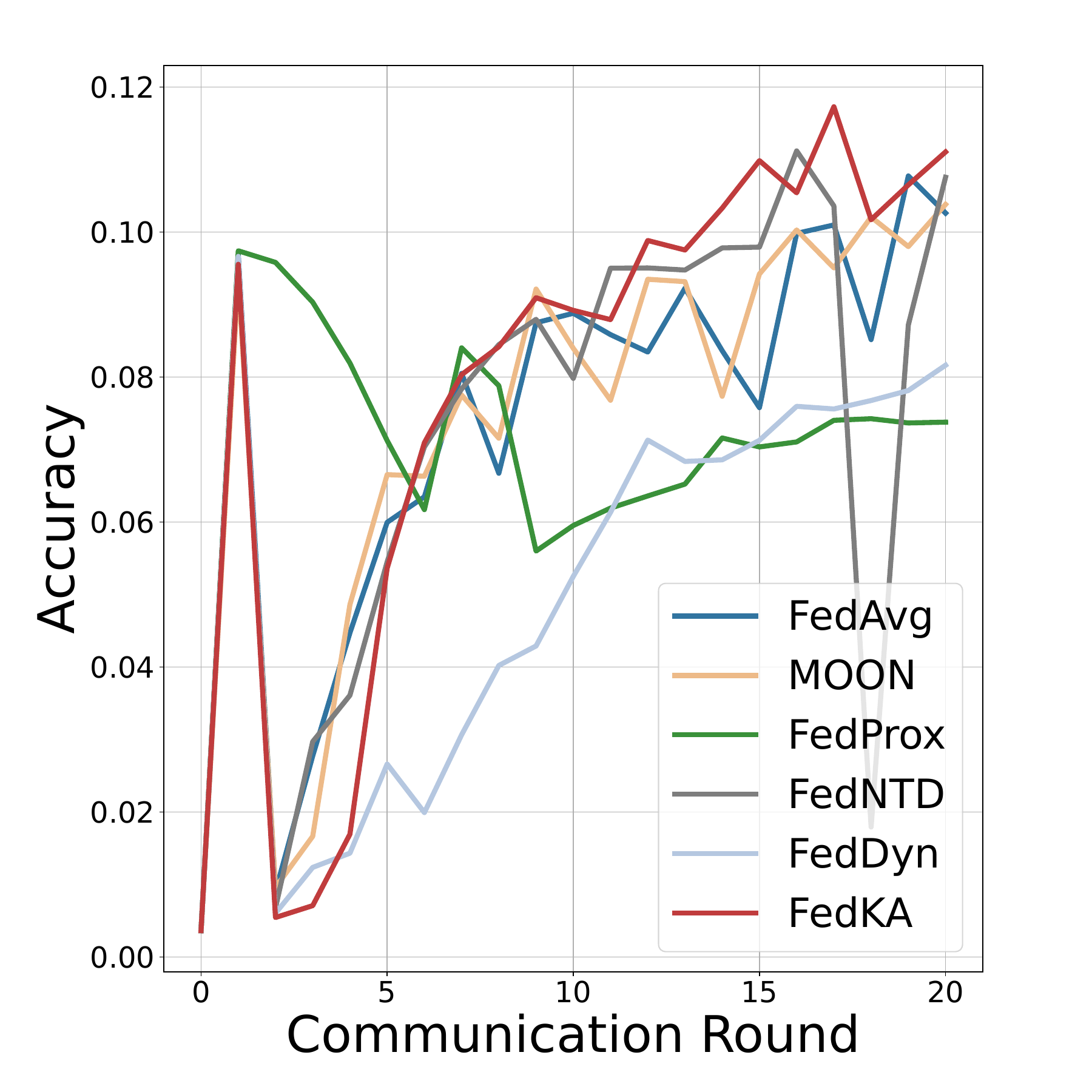}}
    
    \caption{Test accuracy in different communication rounds. FedKA achieves fast and stable convergence with high performance.}
    \label{fig5_1}
\end{figure*}

\subsection{Accuracy Comparison}
We compare the accuracy of the FedKA and other methods in Cifar10, Cifar100, and Tiny-ImageNet datasets with heterogeneous settings. For a fair comparison, we use the same default settings and carefully tune the corresponding hyperparameters in the aforementioned range to get the best performance. The results are displayed in Table \ref{acc_noniid}. Compared with other methods, we can observe that our FedKA achieves great performance on all three datasets. In the Cifar10 dataset, it outperforms the FedAvg by around 2\% accuracy. In the Cifar100 dataset, it also outperforms other methods with around 3\% accuracy on average while slightly lower than FedNTD. The results demonstrate that our proposed federated knowledge anchor can alleviate the impact of heterogeneous data and improves model performance.

\begin{table}[tbp]
  \centering
  \caption{Test accuracy of FedKA and other comparing methods on heterogeneous data.}
    \begin{tabular}{cccc}
    \toprule
    Method & Cifar10 & Cifar100 & Tiny-ImageNet \\
    \midrule
    FedAvg & 0.598 &   0.267  &  0.108 \\
    FedProx & 0.555      &   0.193  &  0.097\\
    MOON &   0.590    &   0.271    &  0.104\\
    FedDyn & 0.571 & 0.252 & 0.097 \\
    FedNTD  & 0.562  &   \textbf{0.299}    &  0.111\\
    FedKA &\textbf{ 0.619} &   {0.293}    &  \textbf{0.117}\\
    \bottomrule
    \end{tabular}%
  \label{acc_noniid}%
\end{table}%

\subsection{Communication Efficiency}
We further compare the communication efficiency of our FedKA with other methods. Since the convergence of each method on Tiny-Imagenet is poor, we only conduct experiments on Cifar10 and Cifar100. We display the accuracy of federated model in each communication round during training in Fig. \ref{fig5_1}. As can be seen, our FedKA converges significantly faster than all the remaining methods. In the Cifar10 dataset, FedNTD initially converges slightly faster than FedKA, but then it hits a bottleneck and fails to achieve higher model performance. However, compared with the other methods, FedKA is more communication-efficient. The accuracy achieved by FedKA at the 40th communication round requires about 100 communication rounds of other methods. 

We further display the number of communication rounds required to achieve the accuracy of FedAvg for 100 communication rounds in Table \ref{table5_2}. As can be seen, FedKA significantly outperforms the remaining methods in terms of communication efficiency. On the heterogeneous Cifar10 dataset ($\alpha = 0.1$), the accuracy of the other methods in training fails to exceed the baseline FedAvg. FedKA, on the other hand, achieves ~ 2.3x speedup, taking only 43 rounds to achieve the performance that FedAvg takes 100 rounds to achieve. Similarly, on the Cifar100 dataset, FedKA also achieves the most significant speedup, achieving 2.7 times speedup. It takes only 36 rounds to achieve the performance of FedAvg for 100 rounds, which is significantly higher than the 2.2 times speedup of FedNTD.

\begin{table}[tbp]
  \centering
  \caption{Communication rounds required by different methods to achieve the accuracy achieved in 100 rounds of FedAvg. The speedup is calculated based on the FedAvg.}
    \begin{tabular}{ccccc}
    \toprule
    \multirow{2}[4]{*}{Method} & \multicolumn{2}{c}{Cifar10} & \multicolumn{2}{c}{Cifar100} \\
\cmidrule{2-5}          & rounds & speedup & rounds & speedup \\
    \midrule
    FedAvg & 100   & 1x    & 100   & 1x \\
    FedProx & $\backslash$  &   $\backslash$    & $\backslash$  & $\backslash$ \\
    MOON  & $\backslash$  &   $\backslash$    & 89    & 1.1x \\
    FedDyn & $\backslash$ & $\backslash$ & $\backslash$ & $\backslash$ \\
    FedNTD & $\backslash$  &    $\backslash$   & 45    & 2.2x \\
    FedKA & \textbf{43}    & \textbf{2.3x}  & \textbf{36}    & \textbf{2.7x} \\
    \bottomrule
    \end{tabular}
  \label{table5_2}%
\end{table}%

\subsection{Robustness to Partial Participation}
As partial participation of clients is a common practice in the real scenario of federated learning, robustness to partial participation is also quite essential metric for federated learning algorithms. Its impact on heterogeneous data is particularly significant, which always leads to jitter in convergence and performance degradation.

Here we show the performance of different methods with various client participation ratios in Table \ref{tabel5_3}. As demonstrated in the table, FedProx has the best performance when the sampling ratio is 0.2, while its accuracy is lower than the baseline FedAvg as the sampling ratio increases. FedKA shows great performance under various sampling ratios, far exceeding the performance of other algorithms, showing strong robustness to the partial participation of clients. 

\begin{table}[tbp]
  \centering
  \caption{Test accuracy of different methods under partial participation strategy on heterogeneous Cifar10. We report the test accuracy under participation ratio: $\{0.2, 0.6, 1.0\}$ to represent low, high, and full participation scenarios.}
    \begin{tabular}{cccc}
    \toprule
    Method & low: 0.2  & high: 0.6  & full: 1.0 \\
    \midrule
    FedAvg & 0.384 & 0.591 & 0.598 \\
    FedProx & 0.480  & 0.549 & 0.555 \\
    MOON  & 0.238 & 0.419 & 0.59 \\
    FedDyn & \textbf{0.482} & 0.548 & 0.571 \\
    FedNTD & 0.398 & 0.588 & 0.562 \\
    FedKA & 0.458 & \textbf{0.591} & \textbf{0.619} \\
    \bottomrule
    \end{tabular}%
  \label{tabel5_3}%
\end{table}%

\subsection{Robustness to Local Epochs}
Due to the deviation between the optimal solutions of the objective function on the local dataset and the global dataset, the number of local epochs plays a crucial role in federated learning. The larger the number of local epochs, the greater the local drift becomes.

We conduct experiments on the Cifar10 dataset with varying local epochs and present the results of different methods in Table \ref{epoch}. The table shows that when the local epoch is small, the difference in performance between the algorithms is not significant. However, as the local epoch increases, the subsequent increase in local drift leads to performance degradation in the other algorithms. In contrast, FedKA maintains a robust performance with significantly higher test accuracy than the other algorithms.

\subsection{Robustness to Client Scalability}
We also conduct experiments to validate the robustness of FedKA to the client scalability. We partition the heterogeneous Cifar10 dataset into 10, 20, 50, and 100 clients respectively. The experimental results are displayed in Table \ref{clientNumber}. Experimental results demonstrate that our FedKA outperforms all other SOTA methods and is robust to client scalability. 

\begin{figure}[t!]
    \centering
    \subfigure[Forgetting on non-dominant classes]{
    \includegraphics[width=0.35\textwidth ]{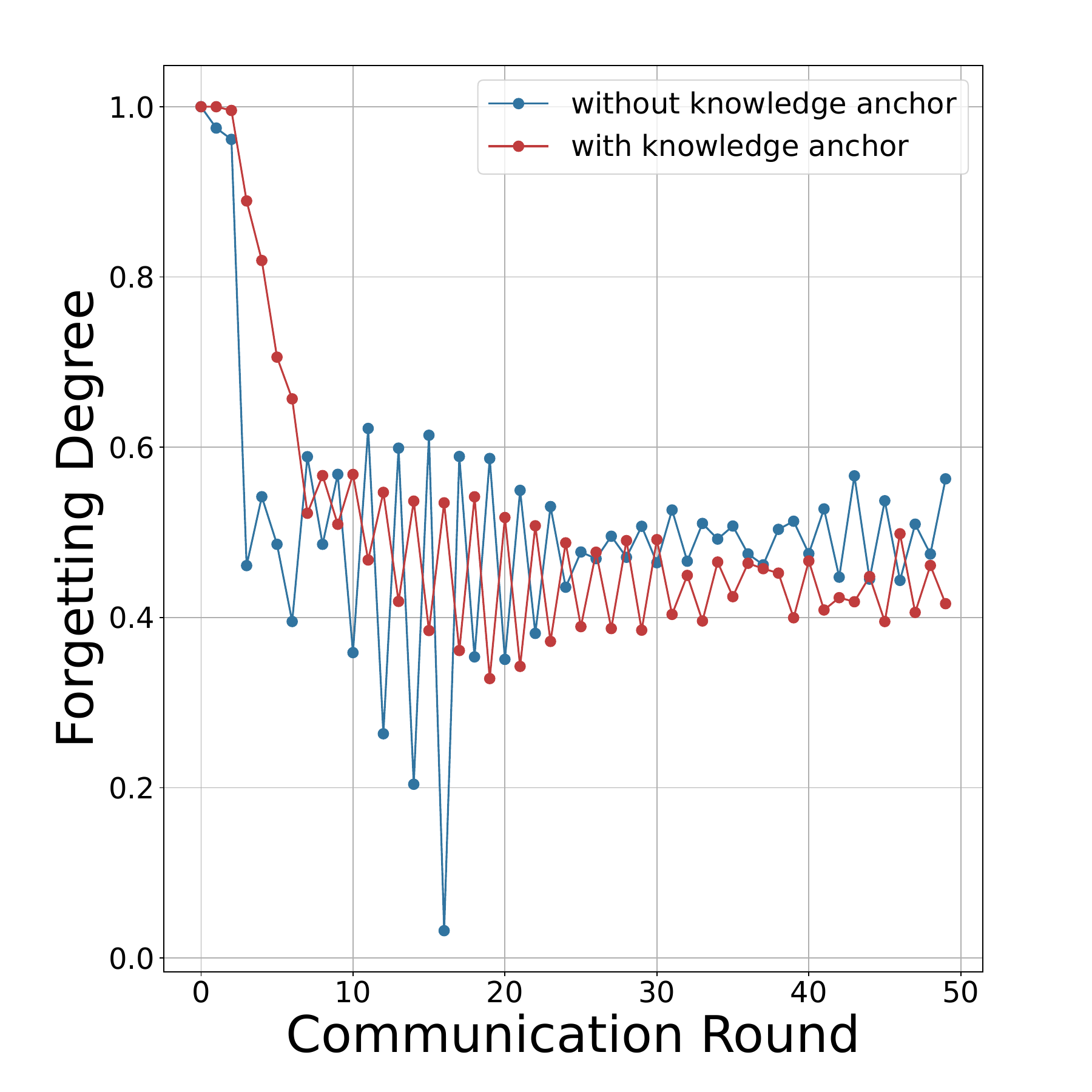}}
    \subfigure[Accuracy on non-dominant classes]{
    \includegraphics[width=0.35\textwidth ]{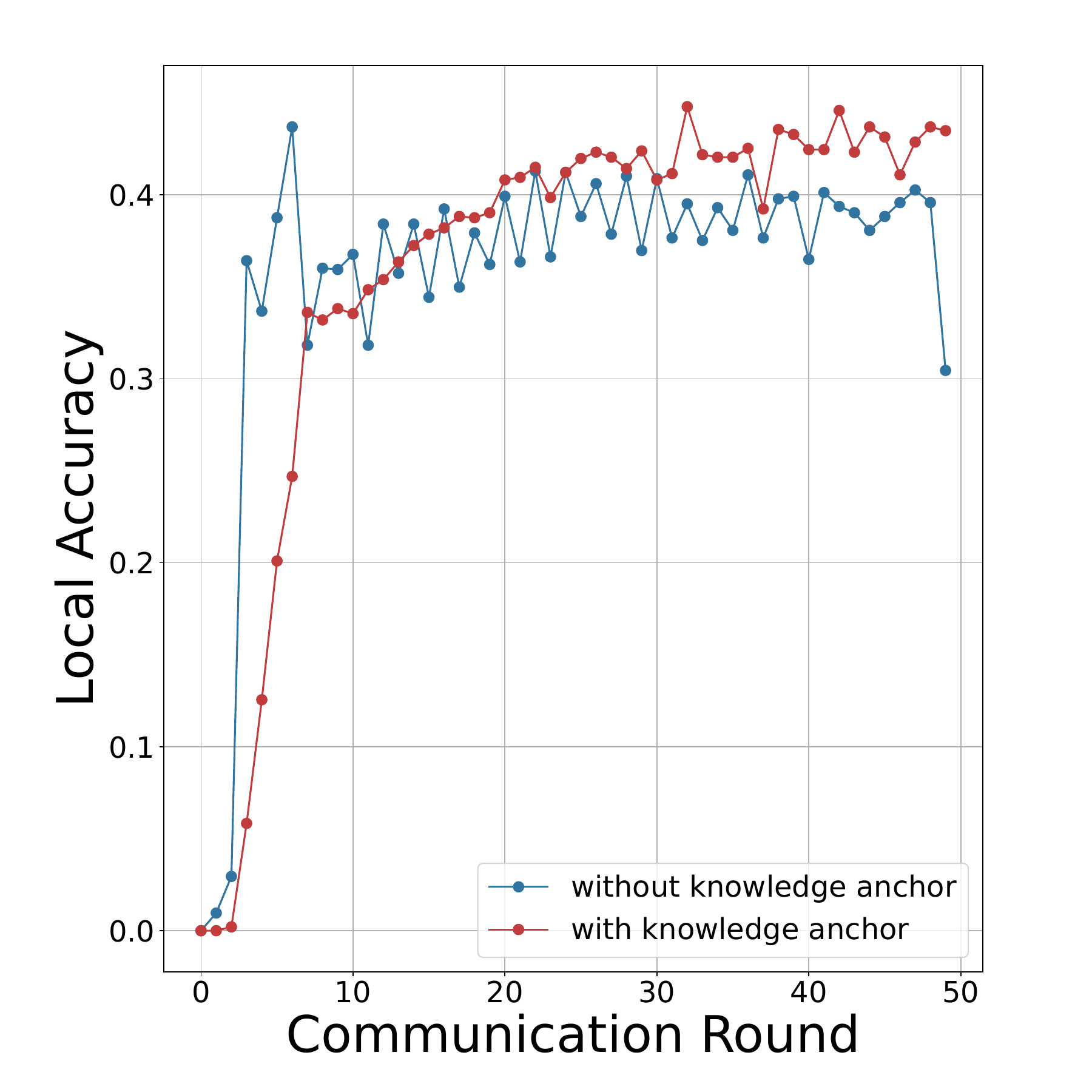}}

    \caption{Forgetting experiments on the non-dominant classes. The knowledge anchor can significantly alleviate the forgetting in non-dominant classes.}
    \label{fig5_2}
\end{figure}

\begin{table}[tbp]
  \centering
  \caption{Test accuracy of different methods under various local epoch numbers.}
    \begin{tabular}{ccccc}
    \toprule
    Method & epoch: 1 & epoch: 5 & epoch: 10 & epoch: 20 \\
    \midrule
    FedAvg & 0.569 & 0.614 & 0.598 & 0.601 \\
    FedProx & 0.407 & 0.529 & 0.555 & 0.567 \\
    MOON  & 0.574 & 0.615 & 0.590  & 0.563 \\
    FedDyn & 0.553 & 0.568 & 0.571 & 0.565 \\
    FedNTD & 0.570  & 0.614 & 0.562 & 0.606 \\
    FedKA & \textbf{0.576} & \textbf{0.617} & \textbf{0.619} & \textbf{0.613} \\
    \bottomrule
    \end{tabular}%
  \label{epoch}%
\end{table}%

\begin{table}[t!]
  \centering
  \caption{Test accuracy of different methods under various client numbers.}
    \begin{tabular}{ccccc}
    \toprule
    Method & Clients: 10 & Clients: 20 & Clients: 50 & Clients: 100 \\
    \midrule
    FedAvg & 0.598 & 0.585 & 0.591 & 0.571 \\
    FedProx & 0.555 & 0.517 & 0.518 & 0.428 \\
    MOON  & 0.59  & 0.567 & 0.578 & 0.544 \\
    FedDyn & 0.571 & 0.551 & 0.572 & 0.524 \\
    FedNTD & 0.562 & 0.588 & 0.593 & 0.525 \\
    FedKA & \textbf{0.619} & \textbf{0.601} & \textbf{0.604} & \textbf{0.573} \\
    \bottomrule
    \end{tabular}%
  \label{clientNumber}%
\end{table}%

\subsection{Ablation Study}
Here we conduct ablation study to demonstrate the effectiveness of our knowledge anchor and its design choice.

\textbf{Forgetting Alleviation.} We conduct experiments to validate the efficiency of the knowledge anchor in knowledge preservation. The proposed forgetting degree serves as the evaluation metric, and we compare the forgetting degree and the local test accuracy of the same class with and without the knowledge anchor. Fig. \ref{fig5_2} displays the forgetting degree curve and the local test accuracy of a specified non-dominant class. The figure illustrates that the forgetting degree of federated learning with knowledge anchor is generally lower than that of federated learning without knowledge anchor, and it is more stable. Additionally, the test accuracy of the federated learning algorithm with knowledge anchor is higher on non-dominant classes, which indicates more preserved knowledge in local training. 

\textbf{Effectiveness of Knowledge Anchor.}
To demonstrate the effectiveness of the construction of knowledge anchor (KA), we perform experiments with different construction methods of knowledge anchors, e.g. without KA, KA with non-dominant classes only, KA with missing classes only, and FedKA. The results are displayed in the Table \ref{KA_1}.

\begin{table}[t!]
  \centering
  \caption{Effectiveness of the knowledge anchor. KA (N) and KA (M) refer to the knowledge anchor with non-dominant classes only and dominant classes only respectively.}
    \begin{tabular}{ccccc}
    \toprule
       Dataset   & Without KA & KA (N) & KA (M) & FedKA \\
    \midrule
    Cifar10 & 0.598 & 0.607 & 0.615 & \textbf{0.619} \\
    Cifar100 & 0.267 & 0.282 & \textbackslash{} & \textbf{0.293} \\
    Tiny-ImageNet & 0.108 & 0.114 & 0.110  & \textbf{0.117} \\
    \bottomrule
    \end{tabular}%
  \label{KA_1}%
\end{table}%

\begin{table}[t!]
  \centering
  \caption{Performance of different sampling strategies in knowledge anchor construction. FedKA (R) refers to the random sampling strategy, FedKA (H) refers to the hard sample mining strategy, FedKA (P) refers to the proficient sample mining strategy.}
    \begin{tabular}{ccccc}
    \toprule
    Strategies & Round 10 & Round 30 & Round 50 & Round 100 \\
    \midrule
    FedKA (R) & 0.513 & 0.591 & 0.601 & 0.619 \\
    FedKA (H) & 0.522 & 0.591 & 0.604 & 0.615 \\
    FedKA (P) & 0.523 & 0.590 & 0.604 & 0.616 \\
    \bottomrule
    \end{tabular}%
  \label{KA_2}%
\end{table}%

\textbf{Knowledge Anchor Construction Strategy.} We also conduct experiments to investigate the influence of sample choosing in the construction of knowledge anchor. We adopt three sample choosing strategies for sampling from non-dominant classes: (1) hard sample mining, (2) proficient sample mining, and (3) random sample selection. The idea of hard sample mining is straightforward, i.e., preserving the knowledge of the hardest non-dominant samples to maximize knowledge preservation. Concretely, before local training, we calculate the sample loss for each non-dominant sample and select the sample with the largest loss as the representative sample for each class. The idea of proficient sample mining is different from hard sample mining. Concretely, the pattern of the hard sample may be false or even meaningless, which cannot represent the current knowledge of the non-dominant class. Conversely, we should use the most proficient sample to serve as the knowledge anchor to preserve the knowledge of non-dominant classes, i.e., the sample with the smallest loss. We adopt these three strategies and conduct experiments on Cifar10 with a heterogeneous setting. The results are displayed in Table \ref{KA_2}. As demonstrated in the results, the influence of the sampling strategy is trivial, and we adopt the random sampling strategy for minimal computation cost.

\section{Conclusion}
Federated learning suffers from forgetting on heterogeneous data. The class-wise forgetting of preserved knowledge happens after local training, leading to performance degradation and blocking the convergence of the federated model. In this paper, building on the previous analysis of \cite{lee2022preservationntd}, we reveal that forgetting only occurs in non-dominant and missing classes. Moreover, we demonstrate that the reduction in samples of non-dominant classes has a trivial impact on forgetting in this class, while the reduction in samples of dominant classes significantly decreases its training benefit and may suddenly lead to catastrophic forgetting when the sample number decreases to a certain threshold. These observations illustrate that local clients struggle to utilize the few samples to combat forgetting in non-dominant classes, motivating us to propose our method, Federated Knowledge Anchor (FedKA). FedKA aims to utilize a minimal representative sample set as an anchor point in knowledge preservation. By minimizing the L2 distance between discarded logits of knowledge anchor outputted by the global model and the current training model, FedKA corrects the gradients towards the direction of preserving the knowledge of non-dominant and dominant classes. Through extensive experiments, we validate that FedKA significantly improves model performance and achieves fast and stable convergence in popular datasets.

\bibliographystyle{unsrt}  
\balance
\bibliography{main}

\newpage
\appendix

\section{Experimental Details}
\begin{figure}[hbtp!]
\centering
\subfigure[$\alpha=0.05$]{\includegraphics[width=0.21\textwidth]{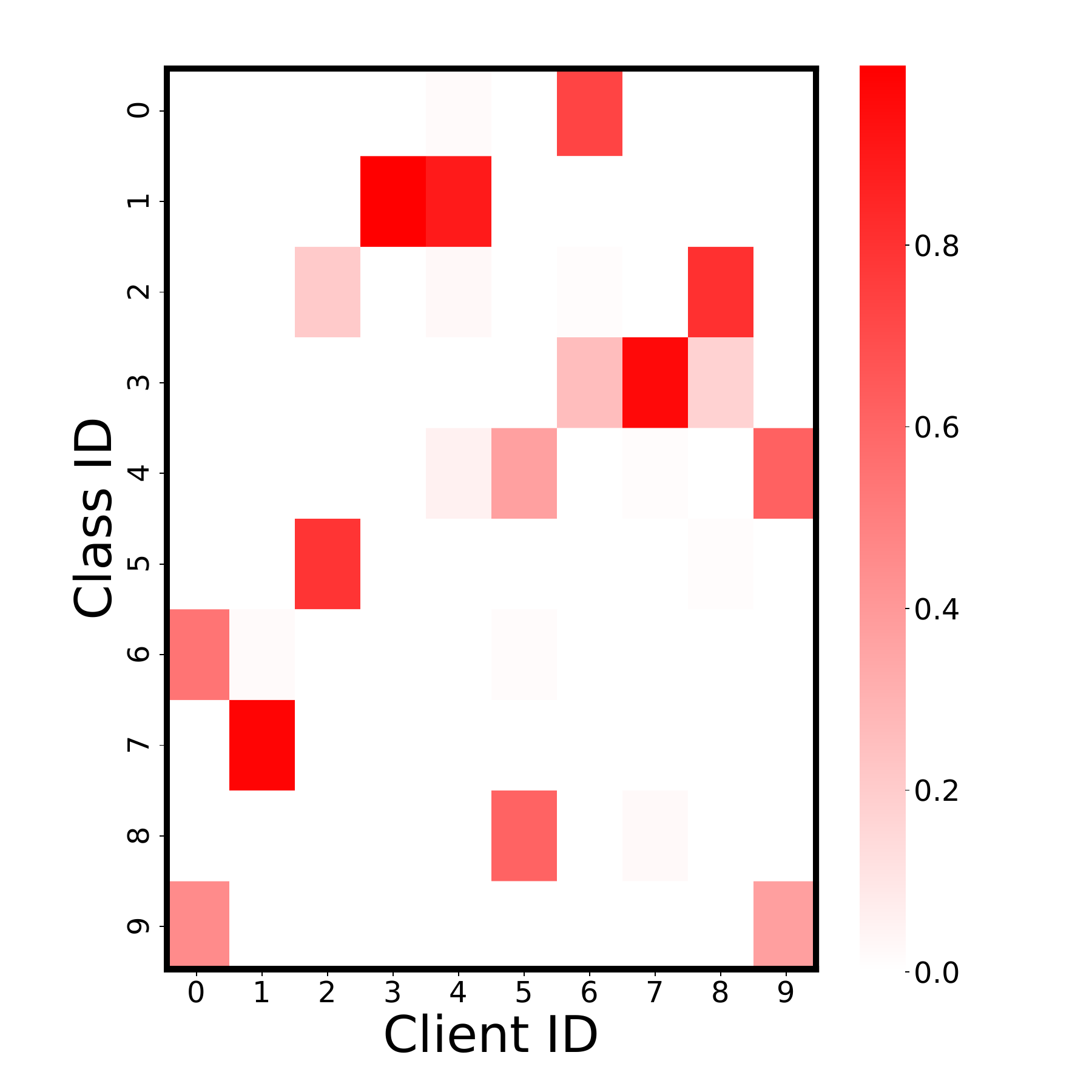}}
\subfigure[$\alpha=0.1$]{\includegraphics[width=0.21\textwidth]{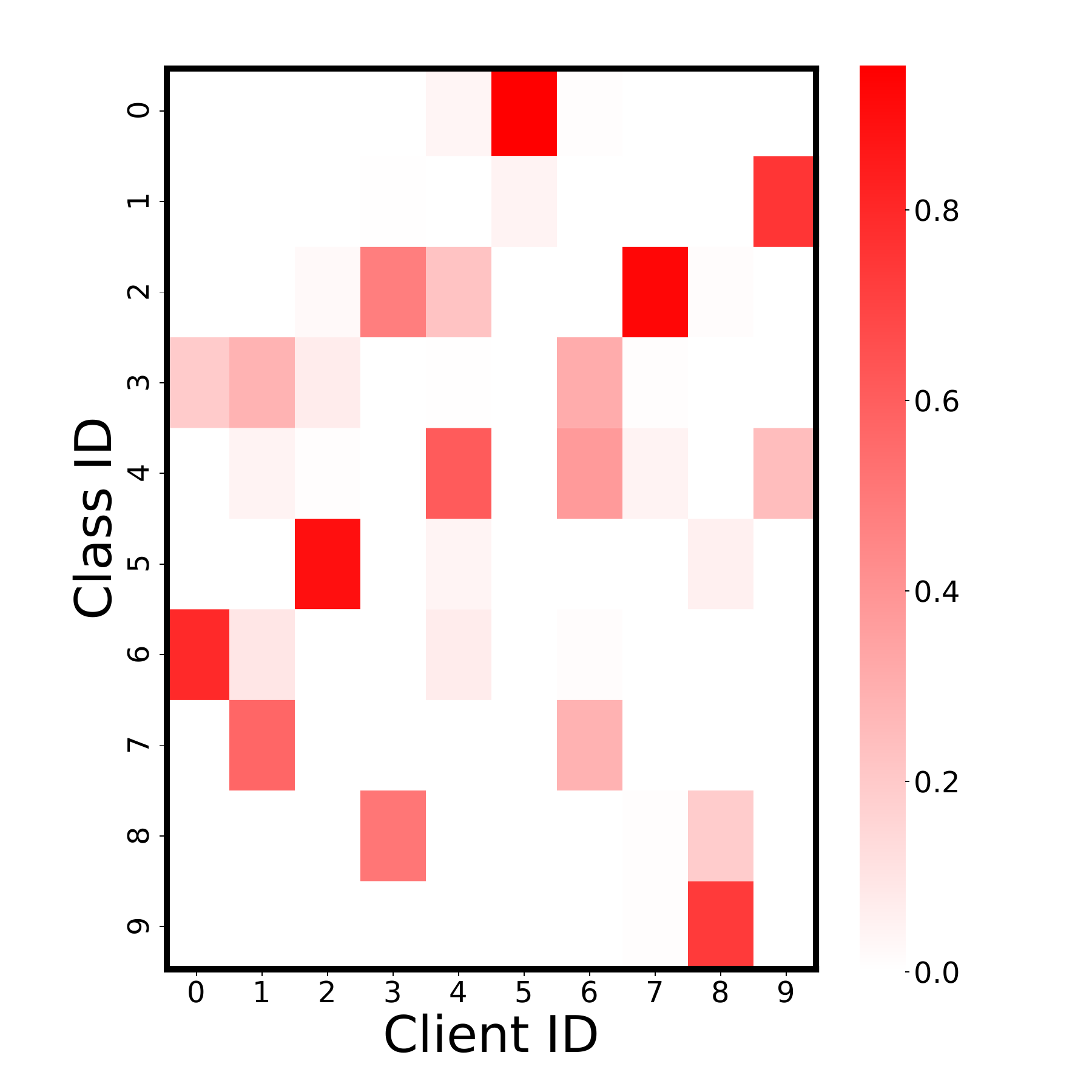}}
\subfigure[$\alpha=0.5$]{\includegraphics[width=0.21\textwidth]{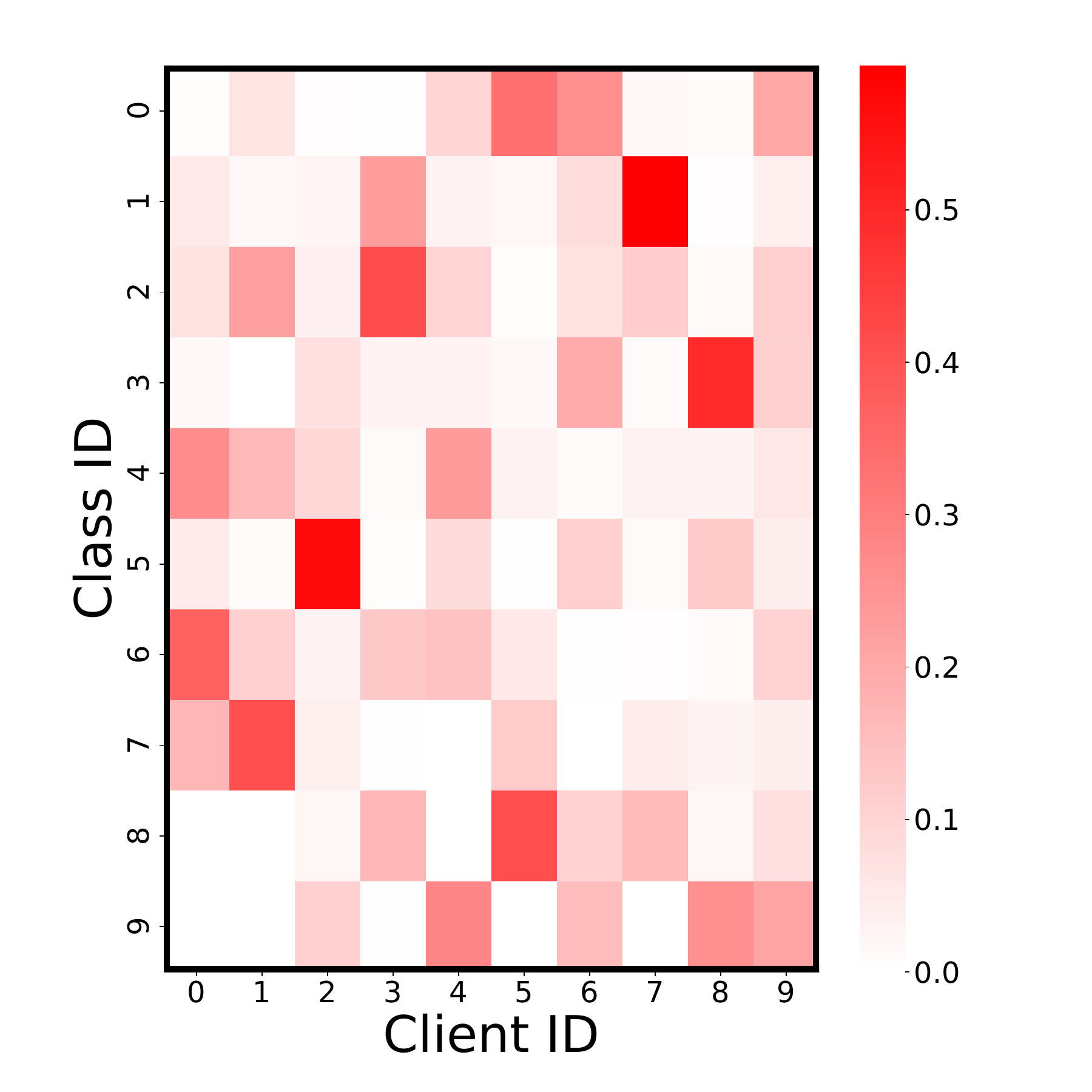}}
\caption{Visualizations of data distribution with varying degrees of heterogeneity severity on the Cifar10 dataset.}
\label{appendix_fig}
\end{figure}
\subsection{Network Architecture}
The t-CNN is a typical convolutional neural network. It consists of two convolutional layers with 32 and 64 filters, respectively, followed by ReLU activation functions and max-pooling operations. The kernel size of the max-pooling layer is 2. The output is then flattened and fed into a fully connected layer with 512 neurons followed by the ReLU activation. The final layer is also a linear layer with the number of neurons corresponding to the specified number of classes.

\subsection{Robustness to Heterogeneity}
We also conduct experiments to demonstrate the robustness of our FedKA to data heterogeneity. Concretely, we set the $\alpha$ of the Dirichlet distribution to $0.05, 0.1$, and $0.5$ respectively, and generate client data with varying degrees of heterogeneity severity on the Cifar10 dataset. We visualize the data distributions in Fig. \ref{appendix_fig}. We evaluate the performance of our FedKA and other SOTA methods. Experimental results are displayed in Table \ref{appendix_table_1}. As demonstrated in the table, our method is robust to severe heterogeneity and outperforms other SOTA methods. 

\begin{table}[htbp]
  \centering
  \caption{Test accuracy of different methods with varying degrees of data heterogeneity severity.}
    \begin{tabular}{cccc}
    \toprule
    Method & $\alpha = 0.05$  & $\alpha = 0.1$   & $\alpha = 0.5$ \\
    \midrule
    FedAvg & 0.578 & 0.598 & 0.601 \\
    FedProx & 0.530  & 0.555 & 0.567 \\
    MOON  & 0.551 & 0.590  & 0.563 \\
    FedNTD & 0.577 & 0.562 & 0.606 \\
    FedKA & \textbf{0.582} & \textbf{0.619} & \textbf{0.613} \\
    \bottomrule
    \end{tabular}%
  \label{appendix_table_1}%
\end{table}%

\subsection{Generic Performance Comparison with Personalized Methods}
We also conduct experiments to compare the generic performance of our FedKA with other SOTA personalized federated learning methods, e.g. FedALA \cite{FedALA}, FedRoD \cite{FedRoD}, FedFOMO \cite{FedFOMO}, and FedPAC \cite{FedPAC}, etc. Following the strategy of \cite{FedRoD}, we measure the generic performance of these methods compared with our FedKA. The experimental results are displayed in Table \ref{appendix_table_2}. The proposed FedKA exhibits stronger generalization performance with higher generic accuracy.

\begin{table}[htbp]
  \centering
  \caption{Generic performance compared with SOTA personalized federated methods.}
    \begin{tabular}{cccc}
    \toprule
    Method & Cifar10 & Cifar 100 & Tiny-ImageNet \\
    \midrule
    FedRoD & 0.540  & 0.194 & 0.079 \\
    FedFOMO & 0.529 & 0.138 & 0.098 \\
    FedPAC & 0.516 & 0.163 & 0.097 \\
    FedALA & 0.545 & 0.205 & 0.097 \\
    FedKA & \textbf{0.619} & \textbf{0.293} & \textbf{0.117} \\
    \bottomrule
    \end{tabular}%
  \label{appendix_table_2}%
\end{table}%

\subsection{Discussion of Minimal Shared Dataset}
In the proposed FedKA, we assume that all clients collaboratively share a minimal dataset that contains only one sample for one class. The number of samples in the minimal shared dataset is $K$. This assumption may raise the concern about suitability. However, such an assumption is easy to be satisfied.
Firstly, for a $K$-classification task, we just require $K$ samples, which is a negligible portion of the entire dataset (e.g., approximately $\frac{1}{6000}$ for Cifar10/100) and easy to be satisfied. Moreover, the number of clients $N$ is usually greater than $K$, ensuring that only a few clients need to contribute one sample. Secondly, even in cases where real-world data is inaccessible, we can adopt generative approaches to generate representative pseudo-data, such as the FedFTG method \cite{FedFTG}. The use of pseudo-data as knowledge anchors will also be explored further in our future work.

\end{document}